# TCLNET: LEARNING TO LOCATE TYPHOON CENTER USING DEEP NEURAL NETWORK

Chao Tan

College of Computer Science and Engineering, Chongqing University of Technology
istvartan@outlook.com

**ABSTRACT**

*The task of typhoon center location plays an important role in typhoon intensity analysis and typhoon path prediction. Conventional typhoon center location algorithms mostly rely on digital image processing and mathematical morphology operation, which achieve limited performance. In this paper, we proposed an efficient fully convolutional end-to-end deep neural network named TCLNet to automatically locate the typhoon center position. We design the network structure carefully so that our TCLNet can achieve remarkable performance base on its lightweight architecture. In addition, we also present a brand new large-scale typhoon center location dataset (TCLD) so that the TCLNet can be trained in a supervised manner. Furthermore, we propose to use a novel TCL+ piecewise loss function to further improve the performance of TCLNet. Extensive experimental results and comparison demonstrate the performance of our model, and our TCLNet achieve a 14.4% increase in accuracy on the basis of a 92.7% reduction in parameters compared with SOTA deep learning based typhoon center location methods.*

## 1. INTRODUCTION

As a very destructive weather system, typhoons have been widely concerned in modern weather forecasting. Timely and accurately determining the location of typhoon center can determine the area affected by the typhoon intuitively, and also provide guidance for the prediction of typhoon intensity and movement path. In the field of weather forecasting, with the gradual development of meteorological satellite, the infrared satellite cloud imagery based typhoon center location methods have been studied for years. Most of these methods use digital image processing and mathematical morphology operation [1-5], or hand-craft feature matching [6] to perform a series of denoising, segmentation, positioning and other operations on the image. Among them, Tian et al. [7] design an adaptive threshold processing method to segment typhoon cloud system by using two value image connecting region labeling algorithm. Hu et al. [8] demonstrate a robust method of locating the typhoon center based on meteorological satellite and microwave scatterometer data. Xu et al. [9]

project page: https://chao-tan.gitee.io/projects/tcl-net/project-page.html

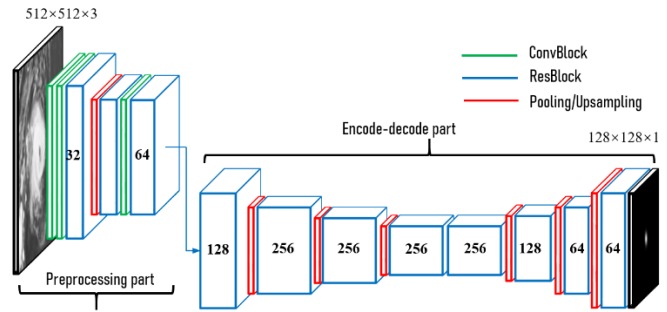

**Figure1.** The overall network structure of TCLNet. It consists of a preprocessing part and encode-decode part stacked by ConvBlocks (green cube), ResBlocks (blue cube) and Pooling/Upsampling layers (red cube). The number of convolution kernels is marked on each ResBlock. TCLNet takes an infrared cloud image of 512×512 size as input and outputs a 128×128 single-channel heatmap.

propose an automatic method to determine the center of tropic cyclones from a serious of SAR images.

However, these algorithms cannot guarantee satisfactory results in complex typhoon scenarios. In recent years, due to the success of machine learning, especially deep learning, has shown that the use of data-driven learning approaches can achieve better results than traditional methods. Therefore, the application of machine learning methods to the field of weather forecasting has become a hot topic in both machine learning and meteorological community. For the task of typhoon center location, Yang et al. [10] present a deep learning based method for typhoon recognition and typhoon center location. Wang et al. [11] propose a two-step scheme deep learning model for locating the tropical cyclone center.

In this paper, we try to use deep learning methods to solve the infrared satellite cloud imagery based typhoon center location problem. We formalize the typhoon center location problem as a 2D keypoints detection problem, and use a deep neural network model for training. Specifically, we design a lightweight fully convolutional end-to-end network including residual blocks, and call this typhoon center location network TCLNet. Meanwhile, in order to train our TCLNet, we also present a brand new large-scale supervised typhoon center location dataset TCLD. To our best knowledge, TCLD is the first large-scale supervised typhoon center location dataset for deep learning research. In addition, we find that manual labels have relative higher errors for complex scenarios such as non-eyed typhoons, and the excessive errors may affect the overall performance of the model. To this end, we propose a novel distance based piecewise loss function named TCL+ loss. The TCL+ loss dynamically reduces the weight for large error samples, and instead learns by using the small error sample, and ultimately improve the overall performance of the model. Experimental results demonstrate that training with the proposed TCLNet with TCL+ loss can effectively reduce the interference of inherent noise in dataset to obtain the best performance. Next, we will present our TCLNet and TCL+ loss function in Chapter 2, and then conduct a lot of experimental comparison and further studies in Chapter 3.

## 2. METHODOLOGY

### 2.1. Typhoon Center Location Formulation

We formulate typhoon center location as a 2D keypoint detection problem. Specifically, we want to learn a deep convolutional network that mapping satellite cloud imagery to typhoon center coordinates. It should be noted that since there is no strict geometric relationship between the typhoon structure and typhoon center, multiple sampling operations will superimpose the prediction error if the coordinate regression is directly performed. Therefore, we use a more robust heatmap regression [12] instead of coordinate regression. To be specific, given an infrared cloud imagery, the network will output a heatmap that is proportional to the input cloud imagery, where the maximum value represents the location of the typhoon center. In addition, we generate the ground-truth heatmaps for each infrared image as follows:

$$H(x,y) = \exp\left(\frac{(x - \alpha \times u)^2 + (y - \alpha \times v)^2}{-2\sigma^2}\right) \quad (1)$$

where $H(x,y)$ is the value of the ground-truth heatmap at $(x,y)$, $(u,v)$ represents the typhoon center coordinate, $\alpha$ and $\sigma$ are the scaling factor and standard deviation respectively, and $\alpha = 1$ when the generated heatmap is the same size as the input infrared image. We will discuss the impact of different heatmap standard deviations on network training in detail in Section 3.3. Therefore, under the setting of heatmap regression, the output of typhoon center location network is a single-channel heatmap with the same size as $H$, and the coordinate of the typhoon center are obtained by detecting the position of the maximum value in the heatmap. It can be seen from Formula 1 that the value range of each point in the heatmap is (0,1], and the maximum value 1 is obtained at the typhoon center position.

### 2.2. Network Structure

In order to solve the problem of typhoon center location in a lightweight and efficient manner, we propose an end-to-end deep neural network named TCLNet. The overall network structure of our TCLNet is shown in Figure 1. It is a standard fully convolutional encoding-decoding structure, and uses max-pooling and upsampling to adjust the range of the receptive fields. The core component of TCLNet is ResBlock, and the reason for using residual network [13] is that it can obtain fewer parameters and better performance by learning residual functions with reference to the layer inputs instead of learning unreferenced functions, compared with traditional convolutional networks. Specifically, as shown in Figure 2, each residual block compresses the input feature maps through 1×1 convolution to reduce the amount of calculation and parameters for the subsequent convolution operations firstly, and then two ConvBlocks and one 1×1 convolution layer are applied for residual features extraction and channel expansion respectively. The final output of residual block is the addition of input and residual feature maps connected by internal skip connection.

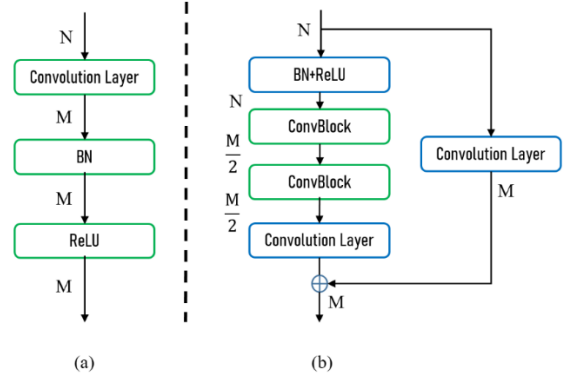

**Figure 2.** The core components (a) convolution block and (b) residual block in TCLNet. BN indicates batch normalization [15] layer and ReLU is Rectified Linear Unit [16]. N and M are the number of input and output feature map channels respectively.

| Layer | Output | Kernel | | | |
|---|---|---|---|---|---|
| Conv-S2 | 16 | 7×7 | ResBlock | 256 | 3×3 |
| ConvBlock | 32 | 1×1 | Maxpooling | 256 | - |
| ResBlock | 32 | 3×3 | ResBlock | 256 | 3×3 |
| Maxpooling | 32 | - | ResBlock | 256 | 3×3 |
| ResBlock | 32 | 3×3 | Upsample | 256 | - |
| Conv | 64 | 1×1 | ResBlock | 128 | 3×3 |
| ResBlock | 64 | 3×3 | Upsample | 128 | - |
| ResBlock | 128 | 3×3 | ResBlock | 64 | 3×3 |
| Maxpooling | 128 | - | Upsample | 64 | - |
| ResBlock | 256 | 3×3 | ResBlock | 64 | 3×3 |
| Maxpooling | 256 | - | ConvBlock | 64 | 1×1 |
| | | | Conv | 1 | 1×1 |

**Table 1.** The architecture for TCLNet. Conv means convolutional layer, ConvBlock means convolution block and ResBlock means residual block. **Output** denotes the amount of output channels in current layer, and **Kernel** means the convolutional kernel size. The convolutional kernel moving stride for all layers is 1 except for the layer name ending with S2 which is 2.

It should be noted that, as with [14], our TCLNet outputs the heatmap of the size only a quarter of the input image, considering that output the heatmap with the same size as the input image of the network will greatly increase the number of model parameters. Therefore, as shown in Figure 1, our TCLNet needs to first downsample the input image to a quarter size, which is accomplished by a preprocessing net of stacks of several ConvBlocks and ResBlocks. Subsequently, TCLNet obtain the final typhoon center heatmap through an encode-decode net of three downsample/upsample operations. The detailed structure of TCLNet is shown in Table 1. It can be seen that our TCLNet follows the simplest network architecture and does not use other strategies like skip connection between encoding and decoding layers or deep supervision. This is because we find that these components do not seems to improve the performance of typhoon center location, even though they are obviously helpful in the task of keypoint detection, and we will discuss in detail in Section 3.3. Meanwhile, we will also study the influence of different network structure settings on our TCLNet in Section 3.3.

### 2.3. Loss Function

In this work, we train the TCLNet using mean square error:

$$L_{mse} = \sum_{1}^{N}(p_i - h_i)^2 \quad (2)$$

where $N$ denotes the total number of pixels of the ground-truth heatmap $h$ and $p$ is the predicted heatmap. Furthermore, we observe that the model has large differences in mean square error (MSE) values for different samples in the course of the experiments, and the loss distribution is long-tailed. We think that this is due to the large errors in the labels of hard samples in the dataset. For further analysis, we divide the testing samples into two categories: eyed typhoon and non-eyed typhoon, which intuitively represent easy samples and hard samples respectively. This is because we find that the loss value of 81.6% of the eyed typhoons is far less than 66.1% of the non-eyed typhoons. In order to test our conjecture, we ask meteorologists to mark the typhoon center of testing samples again, and find that the error of the sample with non-eyed typhoons is about 3.9 times that of the sample with eyed typhoons by comparing with the original labels of the dataset. The above results indicate that because it is more difficult to accurately determine the center of typhoon in complex scenarios such as non-eyed typhoon, the labels of such hard samples often with larger errors than easy samples.

The training dataset contains about 35% non-eyed typhoon samples, and these hard samples contribute a large loss value during the training process, thus affect the model to optimize in the correct direction. To solve this problem, we suppress the loss of hard samples in the training process, and instead learning it through easy samples. Specifically, we propose a new loss function named TCL+ for the task of typhoon center location:

$$L_{tcl+} = \min(L_{mse}, \exp(-2 \times 10^4 \times L_{mse})) \quad (3)$$

The image of TCL+ loss function is shown in Figure 3, where the horizontal axis represents $L_{mse}$ of the sample and the vertical represents the corresponding $L_{tcl+}$ value. It can be seen from the figure that $L_{tcl+} = L_{mse}$ when the $L_{mse}$ of sample is less than about 0.0004, and when $L_{mse}$ is greater than 0.0004, the $L_{tcl+}$ decreases rapidly with the increase of $L_{mse}$ and finally approaches 0. And 0.0004 is the function boundary of $L_{mse}$ between the easy sample and hard sample. It should be noted that, different from directly setting the loss weight of hard sample to 0, since we cannot accurately find the boundary of easy sample and hard sample, a more reasonable solution is to smoothly suppress the loss weight of hard samples with the increase of $L_{mse}$ on the premise that the loss value of hard samples will not exceed that of easy samples. In this way, only when the $L_{mse}$ of the sample deviates too far from the center of loss distribution, its $L_{tcl+}$ will be close to 0. We compare the effects of $L_{mse}$ and $L_{tcl+}$ in Section 3.1, and the result shows that training with our $L_{tcl+}$ can achieve better performance than traditional MSE.

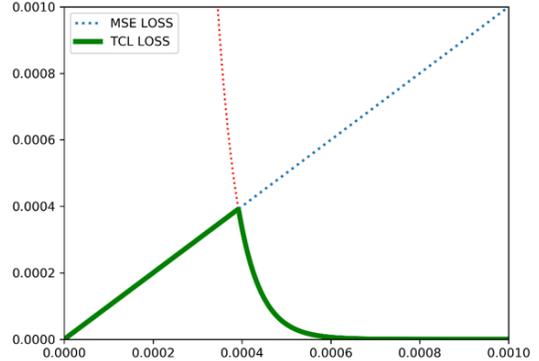

**Figure 3**. The function image of proposed TCL+ loss (solid green line). The horizontal axis represents MSE loss and the vertical represents the corresponding TCL+ loss value.

### 2.4. Implement details

We use the Adam solver [17] with a basic learning rate of 0.001 and the first and second momentum values are 0.5 and 0.999 respectively to optimize our network. To be fair, the size of input infrared cloud imagery is 512×512 for all the experiments, and the size of output typhoon center heatmap is 128×128. In terms of data argumentation, we scale the images from dataset to 574×574 and then randomly cropped to 512×512 with random flips. Unless explicitly specified, we adopt mean square loss (MSE) for all experiments and set the scaling factor and standard deviation in Formula **1** to 0.25 and 15 respectively. We train our network in a total of 65 epochs with a mini-batch size of 4, and we reduce the learning rate from 0.001 to 0.0001 after 30 epochs.

When using TCL+ as the loss function during training, in order to make the network converge faster, we use the mean square error (MSE) loss to train the network at the first 50 epochs and switch to TCL+ loss after 50 epochs to continue training until the end. Lastly, we built our model on PyTorch library and trained

### 3. EXPERIMENTS

| MODEL | MLE-A | MLE-E | MLE-N | PARMS(M) |
|---|---|---|---|---|
| ResNet50 [13] | 51.299±1.0469 | 46.187±0.9739 | 60.747±1.4753 | 23.512 |
| SimpleBaseline-ResNet50 [18] | 6.7906±0.1275 | 4.5178±0.1607 | 10.991±0.3168 | 33.996 |
| SimpleBaseline-ResNet50-up [18] | 6.2601±0.0588 | 5.1693±0.1968 | 8.2760±0.2562 | 29.409 |
| FPN-ResNet50 [19] | 5.5177±0.0440 | 3.7764±0.0572 | 8.7358±0.1329 | 24.965 |
| CPN-ResNet50 [20] | 5.3182±0.0848 | 3.3225±0.1245 | 9.0067±0.1301 | 27.222 |
| CPN-ResNet101 [20] | 5.8058±0.2934 | 3.7922±0.0823 | 9.5274±0.7770 | 46.215 |
| Hourglass Network (×3) [14,10] | 5.2757±0.1490 | 3.5781±0.2143 | 8.4134±0.2130 | 14.998 |
| TCLNet (ours) | **4.5137**±0.0846 | **2.8934**±0.0620 | **7.5083**±0.1684 | **1.0959** |

**Table 2**. Evaluation results on TCLD dataset. **MLE** denotes the mean location error and smaller is better in this metric. **PARMS** means the number of model parameters. Abbreviations: **-A** (all testing samples), **-E** (eyed typhoon samples) and **–N** (non-eyed typhoon samples)

### 3.1. Dataset and Evaluation Metrics

*3.1.1. Typhoon Center Location Dataset (TCLD)*

In order to train our TCLNet, we present a brand new large-scale typhoon center location dataset named TCLD, which contains 1809 training samples and 319 testing samples from Helianthus 8 meteorological satellite. Each sample consists of an infrared satellite cloud imagery that containing only one typhoon and the typhoon center coordinates marked by meteorologist, and the original size of each infrared image is 512×512. Further, in order to observe different performance of the model on eyed typhoon and non-eyed typhoon, we divide the 319 samples in the testing dataset into 207 eyed typhoon samples and 112 non-eyed typhoons. Then, we ask another meteorologist to mark the typhoon center position again and calculate the human error with the testing set labels. Lastly, we calculate that the human errors on eyed typhoons and non-eyed typhoons are 1.68 and 6.55 respectively. We will discuss the evaluation metrics in the next section, and to our best knowledge, TCLD is the first large-scale typhoon center location dataset for deep learning research, which provides rich resources for the research of typhoon center location in the field of machine learning and deep learning.

*3.1.2 Evaluation Metrics*

As with [10], we use the mean location error (MLE) to evaluate our model. MLE calculates the coordinate distance between the predicted typhoon center and typhoon center label on the scale of the input image:

$$MLE = \frac{\sum_{i=1}^{n}\sqrt{(x^i - u^i)^2 + (y^i - v^i)^2}}{n} \quad (4)$$

where $n$ denotes the number of testing samples, $(x, y)$ is the predicted typhoon center, and $(u, v)$ is the label coordinate.

### 3.2. Experiments on TCLD dataset

In this section, we compare TCLNet with many advanced models on the TCLD dataset. It should be noted that since there are few deep learning based typhoon center location methods, we mainly compare with some keypoint detection models. Specifically, we conduct experiments on ResNet [13], SimpleBaseline [18], feature pyramid network (FPN) [19], cascaded pyramid network (CPN) [20] and Hourglass

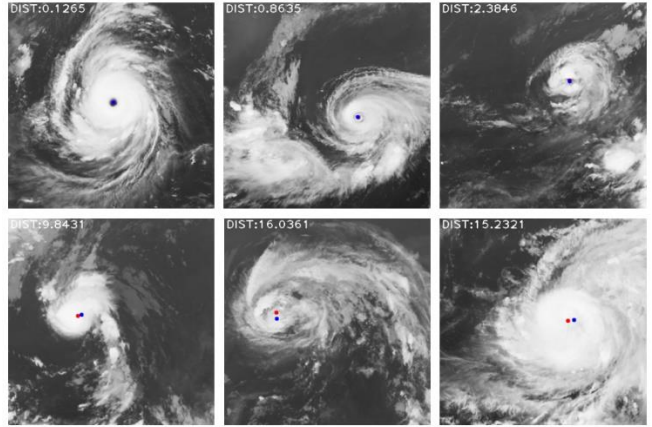

**Figure 4**. Top: three eyed typhoon samples with small errors. Bottom: three non-eyed typhoon samples with large errors. MLE is shown in the upper-left of each image. Blue dots represent the typhoon center labels, and red dots represent the predicted centers.

| MODEL | MLE-A | MLE-E | MLE-N |
|---|---|---|---|
| Human Error | 3.3886 | 1.6800 | 6.5465 |
| TCLNet | 4.5137±0.0846 | 2.8934±0.0620 | 7.5083±0.1684 |
| TCLNet+ | **4.4389**±0.0287 | **2.8462**±0.0305 | **7.3825**±0.1170 |

**Table 3**. Evaluation results on different loss function. The items in the table have the same meaning as in Table 2.

Network [14]. We train the models on TCLD dataset using the official source code and training hyperparameters. In order to make a fair comparison, as with TCLNet, the input image size for all the above models is 512×512, and the size of output typhoon center heatmap is 128×128. In terms of Hourglass Network, we use the same structure setting as [10], that is, three stacked hourglass modules with six scale layers. It should be noted that, except for the ResNet model perform coordinate regression directly, other models use heatmap regression during the training process.

The experimental results are shown in Table 2, in which the model whose name ends with *-up* use a stride 2 deconvolution layer for upsampling, while the other models use nearest neighbor interpolation for upsampling. For each model, in order to eliminate the interference of accidental factors, we carry out 5 repeated experiments and take the mean value of the error as the final experimental result of the model. It can be seen from the last two rows of Table 2 that compare with the current best LocalNet [10], our TCLNet achieves a 14.4% increase in accuracy on the basis of a 92.7% reduction in model parameters. Figure 4 visualizes some test results of TCLNet. It can be seen that TCLNet achieves comparable results on eyed typhoon samples compare with manual labels, and also outputs reasonable results on non-eyed typhoon samples. Next, we will analyze the results of each experiments.

| NO. | HOURGLASS | SCALES | SKIP | MLE-A | MLE-E | MLE-N | PARMS(M) |
|---|---|---|---|---|---|---|---|
| 1 | 3 | 5 | √ | 5.2757±0.1490 | 3.5781±0.2143 | 8.4134±0.2130 | 14.998 |
| 2 | 2 | 5 | √ | 5.2838±0.0873 | 3.4844±0.1201 | 8.4369±0.3202 | 10.099 |
| 3 | 1 | 5 | √ | 5.3050±0.0720 | 3.3696±0.0914 | 9.0366±0.2726 | 5.2012 |
| 4 | 1 | 4 | √ | 5.2895±0.0526 | 3.3031±0.1623 | 8.9597±0.3256 | 4.3602 |
| 5 | 1 | 3 | √ | 5.2547±0.0222 | 3.4891±0.0524 | 8.5178±0.1203 | 3.5192 |
| 6 | 1 | 2 | √ | 5.3316±0.0854 | 3.3937±0.1986 | 8.9133±0.3238 | 2.6783 |
| 7 | 1 | 1 | √ | 6.3172±0.2273 | 3.5399±0.1018 | 11.450±0.6130 | 1.8373 |
| 8 | 1 | 3 | × | **4.9839±0.0281** | **3.2481±0.0962** | **8.1922±0.1617** | **2.1563** |

**Table 4.** Experimental results based on different network structure settings. **NO**. denotes the number of experiment. HOURGLASS means the number of hourglass module. **SKIP** means using skip connection. The other items in the table have the same meaning as in Table 2.

| NO. | MODEL | MULTI | MLE-A | MLE-E | MLE-N | PARMS(M) |
|---|---|---|---|---|---|---|
| 1 | 256×2→256×2→256×2 | × | 5.2099±0.0870 | 3.4524±0.2104 | 8.4581±0.2101 | 3.6580 |
| 2 | 128×2→128×2→128×2 | × | 5.0860±0.0897 | 3.2555±0.1542 | 8.4693±0.1369 | 0.9238 |
| 3 | 256→256→256 | × | 4.9839±0.0281 | 3.2481±0.0962 | 8.1922±0.1617 | 2.1563 |
| 4 | 128→128→128 | × | 4.7929±0.0318 | 3.4635±0.0658 | 7.2500±0.0879 | 0.5456 |
| 5 | 64→64→64 | × | 4.9639±0.0726 | 3.3258±0.0814 | 7.9920±0.3273 | **0.1397** |
| 6 | 64→128→128 | × | 4.6698±0.0602 | 3.1351±0.0845 | 7.5065±0.1424 | 0.3962 |
| 7 | 64→128→128 | √ | 4.7372±0.0323 | 3.0379±0.0464 | 7.8778±0.0460 | 0.6089 |
| 8 | 64→128→256 (TCLNet) | × | **4.5137±0.0846** | **2.8934±0.0620** | **7.5083±0.1684** | 1.0959 |

**Table 5**. Experimental results based on different number of convolution filters settings. **MULTI** denotes use multi-scale deep supervision or not. The other items in the table have the same meaning as in Table 2.

It can be seen from the second row of Table 2 that the error of coordinate based regression model is much larger than that of based on heatmap regression, and this is because the coordinate regression based model loses the fine grained location information during the downsampling process. Further, from the third and fourth rows of the table, it can be seen that the model of using nearest neighbor sampling and convolution layer is better than that of using deconvolution layer, because the deconvolution operation will make the output heatmap contain artifacts, thereby affecting the fitting with label heatmap. In addition, it can be seen from the 6[th] and 7[th] rows of the table that under certain circumstances, more complex feature extraction network will cause performance degradation. We believe this is due to the overfitting caused by the limited number of training data in the TCLD dataset. Lastly, the 8[th] row of the table shows that the network based on the simplest residual architecture is better and more lightweight than the network using the stacked deeper feature extraction module. The experimental results in Table 2 also indicate that it is difficult to effectively improve the performance of the typhoon center location task using strategies such as skip connections between features and deep supervision, and we will discuss in detail in Section 3.3. In general, our TCLNet uses the simplest and lightest design to achieve better performance in the case of the number of parameters are only one-tenth of other models.

In addition, in order to verify the novelty of our TCL+ loss function, we also compare the performance of the mean square error loss (MSE) and our proposed TCL+ loss on the TCLD dataset. The experimental results are shown in Table 3. For the model trained with TCL+ loss, the mean distance error of both eyed typhoon samples and non-eyed typhoon samples is smaller than the model trained with MSE loss. This is because the TCL+ loss suppresses the influence of lager error samples during training the model, so that the model can be trained better.

### 3.3. Further Studies on TCLNet

In this section, we will introduce the process of finding the best model structure and network hyperparameters of TCLNet, and demonstrate that our TCLNet is the simplest and lightweight network structure for typhoon center location task by analyzing the impact of adding different components to TCLNet. It should be noted that the output heatmap size of all models in this section is 128×128, and the training hyper-parameters for different models are all set to the same. At the same time, considering that [10] obtained the state-of-the-art performance by using hourglass network [14] based model before, we analyze various of model structure and network hyper-parameter settings of the hourglass network. The experimental results of different settings are shown in Table 4, and these experiments will provide important guidance for us to design TCLNet. Next, we will briefly introduce the experimental results.

We first study the impact of stacking different numbers of hourglass modules on model performance. It can be seen from the results of experiment No.1 to No.3 in Table 4 that the stacking of multiple hourglass modules enables the performance of the model to obtain a very small gain, but the number of network parameters is multiplied. This is undesirable for the task of typhoon center location, because it means that the model has a greater risk of overfitting and a higher demand of hardware resources. Therefore, we take one hourglass module as the optimal choice. In addition, the depth of the network is also an important consideration in lightweight design. Models that are too deep usually face greater risk of overfitting while achieving good performance, while models that are too shallow cannot achieve good performance for their weak fitting ability. In order to explore the most suitable feature extraction scales for the typhoon center location problem, we conduct experiments on 5 different downsampling scales on the basis of single hourglass network, and the results are shown in the experiments No.3 to No.7 in Table 4. It can be seen that as the scale of feature extraction increases, the mean location error (MLE) of the network first decreases and then increases, and reaches the optimal when the downsampling scale is 3. Our TCLNet also follows the setting of downsampling scales to 3. As for the reason why the performance of the network decreases after the downsampling scale is larger than 3, we believe that in addition to overfitting, another reason is that the excessively small feature map at the bottom of the model leads to deviation of the fine position during the upsampling process, thus resulting in the error of the output heatmap. Lastly, we verify the influence of skip connections between each feature layers for the hourglass network. It can be seen from the results of experiment No.1 and No. 8 in Table 4 that disabling the skip connection between the feature layers in encode-decode part can not only improve the performance of the model, but also reduce the amount of network parameters considerably. Therefore, our TCLNet does not use such skip connections, at the same time, [18] also use similar idea to ours. To sum up, our TCLNet is a single model with 3 downsampling scales and without the use of skip connections between residual blocks.

Next, we will determine the optimal number of convolution kernels for the three feature extraction residual blocks in encode-decode part of TCLNet. Before that, we first compare the impact of different size of convolution kernels on model performance. Noted that inspired by the VGG network [21], we use a stack of two 3×3 convolution layers to replace a larger convolution kernel size layer. It can be seen from the experiments No. 1 and No.3 or No.2 and No.4 in Table 5 that increasing the convolution kernel size or stacking feature extraction layers at same scale will deduce the performance of the model. In terms of the number of filters, we try a total of 5 different setting, where the number of filters for each residual block is one of 64, 128 and 256, and the number of filters in the shallow layer is less than or equal to the number of deeper layer. The experimental results of No.3 to No.6 and No.8 in Table 5 show that increase the number of convolution filters from 64 to 256 gradually can achieve the best balance in terms of the number of parameters and performance. Finally, considering that many existing methods use multi-scale deep supervision during training [19,20], we also explore the necessary of this strategy in typhoon center location task. In experiment No.7 of Table 5, we let each layer in decoding network output the corresponding predicted heatmap through an additional convolution layer, and calculate the loss with label heatmap during training process. A comparison of experiment No.7 and No.8 in Table 5 shows that the use of multi-scale deep supervision will not improve the performance, so we do not use this strategy in TCLNet.

| STD | MLE-A | MLE-E | MLE-N |
|---|---|---|---|
| 5 | 5.1253±0.0741 | 3.4311±0.1437 | 8.2564±0.3290 |
| 10 | 4.6345±0.0417 | 3.0714±0.1158 | 7.5235±0.1912 |
| 15 | **4.5137**±0.0846 | **2.8934**±0.0620 | **7.5083**±0.1684 |
| 20 | 4.7810±0.1145 | 3.0845±0.0616 | 7.9165±0.3020 |
| 25 | 4.9612±0.0391 | 3.2148±0.0640 | 8.1890±0.1735 |
| 30 | 5.7168±0.1376 | 4.2104±0.2082 | 8.5009±0.2371 |

**Table 6**. Experimental results of different standard deviations used to generate heatmap labels. **STD** means the standard deviations.

In the end, we use the model corresponding to experiment No.8 in Table 5 as the optimal model network for typhoon center location. It uses relatively low computational cost to achieve the best performance in all experiments, which is also the network structure of our proposed TCLNet.

During the training process, we also analyze the impact of different standard deviations used to generate the ground-truth heatmap on the training result. We conduct experiments on TCLNet with standard deviations ranging from 5 to 30 in interval of 5, and note that except for the different standard deviations used to generate heatmap labels, the other training parameters are all the same settings. The result of six experiments are shown in Table 6. It can be seen that too small standard deviation will lead to too steep probability value distribution of heatmap, which results in performance degradation due to too strict restrictions on the output of the model. Conversely, a too large standard deviation will increase the supervision error during the training process since the probability value distribution of heatmap is too flat. As can be seen from Table 6 that the best effect can be achieved when the standard deviation is set to 15.

## 4. CONCIUSIONS

In this work, we propose a lightweight fully convolutional network TCLNet and a novel TCL+ loss to solve the problem of typhoon center location. Meanwhile, we also present a large-scale typhoon center location dataset TCLD for deep learning research. Extensive quantitative and qualitative experiments demonstrate that our TCLNet has obvious advantages over SOTA methods in terms of performance and parameter quantity. In the future, we plan to use TCLNet to explore more typhoon based problems, such as typhoon moving path prediction.